# Simulation based Hardness Evaluation of a Multi-Objective Genetic Algorithm

Shahab U. Ansari and Sameen Mansha
Faculty of Computer Science and Engineering
Ghulam Ishaq Khan Institute of Engineering and Science and Technology
Topi, Pakistan
e-mail: sansari@giki.edu.pk, gcs1203@giki.edu.pk

*Abstract—Studies have shown that multi-objective optimization problems are hard problems. Such problems either require longer time to converge to an optimum solution, or may not converge at all. Recently some researchers have claimed that real culprit for increasing the hardness of multi-objective problems are not the number of objectives themselves rather it is the increased size of solution set, incompatibility of solutions, and high probability of finding suboptimal solution due to increased number of local maxima. In this work, we have setup a simple framework for the evaluation of hardness of multi-objective genetic algorithms (MOGA). The algorithm is designed for a pray-predator game where a player is to improve its lifespan, challenging level and usability of the game arena through number of generations. A rigorous set of experiments are performed for quantifying the hardness in terms of evolution for increasing number of objective functions. In genetic algorithm, crossover and mutation with equal probability are applied to create offspring in each generation. First, each objective function is maximized individually by ranking the competing players on the basis of the fitness (cost) function, and then a multi-objective cost function (sum of individual cost functions) is maximized with ranking, and also without ranking where dominated solutions are also allowed to evolve.*

## I. INTRODUCTION

Multi-Objective optimization is the task of searching for optimum solution(s) while optimization problem converges by taking into account multiple objective functions. In real life problems it is usually advantageous to put in many objectives. A subset of the set that contains all reasonable points of solutions optimizing at least one objective while holding further objectives invariable."Evolutionary algorithms" is the phrase used to represent a set of stochastic optimization methods that replicate the method of expected evolution. In the late 1950s the birth of EAs can be traced back. Numerous evolutionary methodologies have been proposed, chiefly genetic algorithms, evolution strategies, and evolutionary programming since the 1970s [1].

Numerous techniques have been developed in order to spawn a complete Pareto set of multi-objective problem through a single run by utilizing the population of an evolutionary algorithm. These methods can be categorized into Pareto based and non-Pareto based methods. Without constructing a straight relationship to verify supremacy/non- supremacy with other elements of the population the non-Pareto based methods generate a Pareto set [2]. By grading the population depending on a straight evaluation of Pareto supremacy within the population Pareto based methods (MOGA, SPEA, NSGA) have been used. A set of candidate solutions is spawned in all of these methods. Selection and variation are basic principles to modify the set of candidate solutions. Selection imitates the opposition for resources among living beings and reproduction, while variation, imitates the expected potential of creating "novel" living beings by applying mutation and recombination.

It can be obstructive or beneficial to add objectives in an optimization problem. Search behavior of evolutionary algorithms and the structure of a given problem are affected by adding objectives in an optimization problem [3]. Due to the change in structure of a given optimization problem running time also varies. Whenever multiple objectives interfere with each other, plateaus of comparable or incomparable solutions may appear or fade away. Adding multiple objectives can decrease the complexity of a problem. By considering different categories of hardness, multiple objectives can facilitate with the organization and design of multi-objective benchmark problems [4].

There is some verification in the literature that multiple objectives can make a problem harder. Many researchers, e.g., in [5],[6],[7] support the assumption that the optimization becomes harder as the more objectives are added. Furthermore, the behavior of a multi-objective evolutionary algorithm on a problem with higher number of objectives cannot be generalized to a few objectives [6].It is proved that that a single-objective problem can be solved more efficiently via a generalized multi-objective model by investigating the single shortest paths problem and the computation of minimum spanning trees[8]. Single objective problems are decomposed into multi-objective problems introducing the concept of "multi-objectivized" making them easier to solve than the original problems[9],[10].

The abilities to accept new information from the environment and use it to update our existing knowledge thus adapting to the changes of our environment have played a crucial role in the success of human beings as a species [11]. Over the period of time computer games have become a major source of entertainment for humans. From the point of view of game developers there is a constant demand of writing games which are entertaining for the end users but entertainment itself is of subjective nature [12]. It has always been difficult to quantify the entertainment value of the human player [13], [14].

Computational intelligence (CI) can help designing games that will be played without human interaction [15]. In this paper a game play area and an agent is defined. Genetic algorithm is used to enable artificial agent to detect multiple changes in environment and adapt accordingly while moving in the game play area. These changes are detected by the controller while considering multiple objectives.

## II. RELATED WORK

There must be some rules to evolve a game. A number of axioms should be laid down, or in other words different assumptions should be made, to decide a rule space. In [15] a game-play area and an artificial agent is defined. A controller instead of a human being is controlling an artificial agent. That controller is base on artificial neural network and is trained by using evolutionary algorithms. Artificial agent is being enabled by controller to sense multiple changes in surroundings. Artificial agent continues moving in game play area while keep adapting different changes. Controller considers objective function and detects changes in environment accordingly.

The game takes place on a discrete grid with dimensions $15 \times 15$ cells are placed on grid. Walls are laid out to demonstrate that rules are evolved not the environment. Each cell on grid is either a wall or free space. This layout can be seen in figure 1. Game will be played for a specific number of time steps, starting at t=0 and continuing until either score of an agent >= scoremax, t = tmax or the death of agent.
If, at the end of a game, score >= scoremax the game is won; otherwise, the game is lost.
At the start of the game, a random position is assigned to agent except walls. High priority is assigned to 4×4 central most cells as compared to other cells. The agent will move one step either left, right up or down at every time step. Any move that would result in the agent occupying the same cell as a wall is not executed. Predators of red, green or blue colors are defined on random and empty positions on grid except walls. Predators may move one step either left, right up or down at every time step. Predators cannot pass through walls.

Movement logic is defined according to which predator moves on grid. Keeping in view collision logic score is updated. According to collision table, it is determined that what happens to predator when two predators of the same or different classes collide, or when a predator and the agent collide. According to score logic, score is updated by considering collision between two predators or between a predator and the agent occurs. Rule space consists of eight simple parameters: maximum score, maximum time, the number of red, blue and green predators, the movement logic for red, green and blue predators, and two tables: collision score effects.

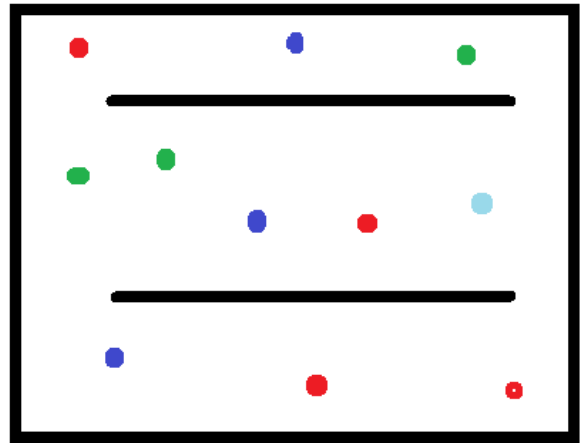

Figure 1: Search Space

Fitness function for game rules based on learning process is calculated each time and game is evolved on the basis of this function. It is a very promising idea to produce new games by evolving game rules. An experiment for evolution of a simple computer game based on [15] is presented in [12,13, 16-19].In this process they have presented some metrics for entertainment which are based on duration of play, level of challenge, diversity in artifact's behavior, and usability of play area. These matrices are combined in a fitness function to guide the search for evolving the rules of the game. The result of their experiment shows that games can be evolved in this manner. It is not proved that the most entertaining game will be the best evolved game as well. By evolution of games, multiple equally interested games can be introduced rather than randomly generated games.

## III. METHODOLOGY

### A. Game Strategy

In this study, an evolution strategy based pray-predator game is developed in which performance of the player is evolved in terms of its lifespan, level of challenge and usability of the game area through a number of generations. Each game is called a chromosome which comprises of 30 genes. Figure 2 shows the allele of each gene of the chromosomes participating in the game. Each game has 100 steps and played for 10 times to get the average value of the cost function. The total number of chromosomes used in the game is 20, which are crossed over and mutated to generate off springs. The number of generation is preset to 100. At the start of the game, the player and predators are placed at random locations on a two dimensional grid of 14-by-14. Two walls of length 7 are also placed in the arena and to be avoided by the player and the predators. The predators are of three types, and their numbers are randomly selected from 0 to 20. The movement strategy of the player is defined as follows: look for a high scoring predator in one of the four neighborhood locations – east, north, west and south. If it is not there then look for an empty location. The predators have to move straight until they reach a wall where they make either a right or left turn based on the value of the genes. The player and predators are also awarded score according to the genes values set randomly at the beginning of the game. The game continues until player evolves to the last generation, the player acquires the maximum score, or the player dies. Figure 3 shows the game algorithm.

| Red, Green, Blue | 0-20 | Number of predators |
|---|---|---|
| Red, Green, Blue | 0-3 | Movement Logic |
| Red-Red | | |
| Red-Green | | |
| Red-Blue | | |
| Red-Agent | | |
| Green-Red | | |
| Green-Green | | |
| Green- Blue | 0-2 | Collision Logic |
| Green-Agent | | |
| Blue- Red | | |
| Blue- Green | | |
| Blue- Blue | | |
| Blue- Agent | | |
| Agent- Red | | |
| Agent-Green | | |
| Agent-Blue | | |
| Red-Red | | |
| Green-Green | | |
| Blue-Blue | | |
| Agent-Red | | |
| Agent-Green | -1,0,1 | Score Logic |
| Agent-Blue | | |
| Green-Red | | |
| Blue-Red | | |
| Blue-Green | | |

Figure 2: Game Logic

1.  Begin program
2   Create chromosomes
3.  Create grid with walls
4.  Place player and predators
5.  Loop for hundred generation
6.  Crossover and mutation
7.  Loop for ten games
8.      Loop for hundred steps
9.          Move player and predators
10.         Update score
11.         Update usability
12.     End loop
13.     Update life
14.  End loop
15.  Compute average life
16.  Compute average usability
17.  Compute challenge level
18.  Rank chromosomes
19. End loop
20. End program

Figure 3: Game Algorithm

*B. Cost Function*

The single-objective algorithm uses life span of the player as the cost function which is maximized through generations. In the algorithm, the life span $L$ is the number of iterations $n$ the player survive the game averaged over total number of games $N$ played as,

$$L = (\textstyle\sum n)/N \qquad (1)$$

In multi-objective algorithm, three objectives are used – player's life span, it's challenge level and it's usability of the game. The challenge level $C$ is modeled as Gaussian function with mean score $\mu$ and standard deviation $\sigma$ as,

$$C = e^{-\frac{1}{2}(\frac{x-\mu}{\sigma})^2} \qquad (2)$$

The usability of the game arena is the average number of cells $c$ a player has visited during $N$ games, and computed as,

$$U = (\textstyle\sum c)/N \qquad (3)$$

The multi-objective cost function is the sum of *L, C* and *U*, that is maximized through generations.

## IV. CURRENT WORK AND RESULTS

### A. EXPERIMENT 1

Experiment 1 is conducted for single-objective genetic algorithm in which life span of the player's life is maximized. The player plays a game of one thousand steps for ten times. Its life is computed using Equation 1 in the previous section. The competent games are ranked according to their life span and are used to produce off springs for the next generation. The number of generation used is 500. The results of the experiment are shown in Table 1.

### B. Experiment 2

Experiment 2 is conducted for the evolution of game area usability. The usability of the game area is the count of the number of cells visited by either the player or the predators. It is computed using Equation 2 in the previous section. The competent games are ranked according to their usability, and are used to produce off springs for the next generation. The number of generation used is 500. The results of the experiment are shown in Table 2.

Table 1 – Results for Player's Life

| Experiment # | Convergence Time |
|---|---|
| 1 | 157 Iterations |
| 2 | Not Converging |
| 3 | 466 Iterations |
| 4 | Not Converging |
| 5 | 1 Iteration |
| 6 | 1 Iteration |

Table 2 – Results for Game Area Usability

| Experiment # | Convergence Time |
|---|---|
| 1 | 1 Iteration |
| 2 | 1 Iteration |
| 3 | 1 Iteration |
| 4 | 1 Iteration |
| 5 | 1 Iteration |
| 6 | 1 Iteration |

Table 3 – Results of MOGA

| Experiment # | Convergence Time |
|---|---|
| 1 | Not Converging |
| 2 | 1 Iteration |
| 3 | 1 Iteration |
| 4 | Not Converging |
| 5 | Not Converging |
| 6 | Not Converging |

### C. Experiment 3

In experiment 3, multi-objective cost function is employed to evaluate the hardness of multi-objective genetic algorithms in terms of convergence and evolution. Three objectives used in the experiment are the player's life span, its challenge level and its usability of the game area. The chromosomes are ranked on basis of the sum of all the three objectives. Only top 10 chromosomes are used in crossover and mutation to produce off springs in each generation. Table 3 shows the outcome of this experiment.

### V. DISCUSSION

In this project, the hardness of the evolution of multi-objective genetic algorithms is studied. Two set experiments are conducted for single-objective and multi-objective genetic algorithms designed to play a pray-predator game. The moving player looks for high scoring pray and the same time avoid any harm to his life in thousand steps. Twenty games are played for ten times to get an average value of the cost functions defined in method section. For single-objective genetic algorithm, two games are played for optimizing life span and three games are played for usability optimization. Life span shows convergence above 400 iterations. Usability, however, does not converge for two games and in the third game it converges right at the beginning of the game. The dependency of the usability on life span of the player is also observed. In multi-objective genetic algorithm, seven experiments are conducted. In four of the games the algorithm does not converge at all. In two games, it converges in two iterations and in one it converges after 69 iterations. The results show that multi-objective genetic algorithms are hard to evolve. However, to generalize this theory, more objectives with more number of experiments should be studied. Another interesting phenomena observed in the study is the pulling property of the dominating solution in which a winner tries to pull others to the best solutions.

## VI. CONCLUSION AND FUTURE WORK

The multi-objective genetic algorithms are hard to evolve as compared to single-object genetic algorithms. This hardness in evolution is due to multi-modality of the cost functions, incompatible solutions and larger solution space. The optimal solution of multi-objective optimization problem always comes from Pareto front, which compromises the optimal solutions by trading them of among all the objectives. To generalize this finding, more number of objectives, up to 20, needs to be tested.